\documentclass[runningheads]{llncs}
\usepackage[T1]{fontenc}
\usepackage{graphicx}
\usepackage{booktabs}
\usepackage{xcolor}
\usepackage{array}
\usepackage{CJKutf8}
\begin{document}
\title{Exploring the Role of Knowledge Graph-Based RAG in Japanese Medical Question Answering with Small-Scale LLMs}
%
\titlerunning{Exploring the Role of Knowledge Graph-Based RAG}
%
\author{
Yingjian Chen\inst{1,*}\orcidID{0009-0004-1359-6998} \and
Feiyang Li\inst{1,*}\orcidID{0009-0007-2982-8206} \and
\\
Xingyu Song\inst{1}\orcidID{0009-0007-1640-0809} \and
Tianxiao Li\inst{2}\orcidID{0000-0002-9147-7511} \and
\\
Zixin Xu\inst{3}\orcidID{0009-0009-1838-4229} \and
Xiujie Chen\inst{4}\orcidID{0009-0006-8111-3307} \and
\\
Issey Sukeda\inst{1}\orcidID{0000-0002-9147-7511} \and
Irene Li\inst{1}\orcidID{0000-0002-1851-5390}}
\authorrunning{Yingjian Chen, Feiyang Li, et al.}
%
\institute{The University of Tokyo, Tokyo, Japan \and
NEC Laboratories America, Palo Alto, USA \and
Dokkyo Medical University, Tochigi, Japan \and
Graduate School of Frontier Sciences, The University of Tokyo, Tokyo, Japan
\\
$^*$ Equal Contribution \\
\email{irene.li@weblab.t.u-tokyo.ac.jp}\\
}
\maketitle              
\begin{abstract}
Large language models (LLMs) perform well in medical QA, but their effectiveness in Japanese contexts is limited due to privacy constraints that prevent the use of commercial models like GPT-4 in clinical settings. As a result, recent efforts focus on instruction-tuning open-source LLMs, though the potential of combining them with retrieval-augmented generation (RAG) remains underexplored.
To bridge this gap, we are the first to explore a knowledge graph-based (KG) RAG framework for Japanese medical QA small-scale open-source LLMs. Experimental results show that KG-based RAG has only a limited impact on Japanese medical QA using small-scale open-source LLMs. Further case studies reveal that the effectiveness of the RAG is sensitive to the quality and relevance of the external retrieved content. These findings offer valuable insights into the challenges and potential of applying RAG in Japanese medical QA, while also serving as a reference for other low-resource languages.

\end{abstract}
\section{Introduction}
\vspace{-2mm}
Large language models (LLMs) have achieved remarkable performance in medical question answering (QA), even demonstrating the ability to pass medical licensing exams (e.g., the United States Medical Licensing Examination, USMLE)~\cite{nori2023capabilities}, which highlights their potential to understand complex medical knowledge. In particular, recent research~\cite{shi2023mkrag,xiong-etal-2024-benchmarking,yang2024kg} has explored the use of retrieval-augmented generation (RAG)~\cite{edge2024local} to incorporate external medical knowledge into LLMs, effectively mitigating ``hallucination'' issues~\cite{mckenna2023sources,yang2025retrieval,zhang2023language} and further enhancing their applicability in medical QA tasks. 


While these advances are promising, most existing efforts~\cite{ke2024mitigating,yang2024llm} focus on English-only scenarios, with limited exploration of multilingual medical QA, particularly in Japanese. Since mainstream LLMs~\cite{hurst2024gpt,touvron2023llama} are predominantly trained on English-centric medical data, the significant imbalance in resource distribution~\cite{chataigner2024multilingual} limits their effectiveness in Japanese medical QA scenarios.
While GPT-4 has demonstrated strong performance on the Japanese NMLE~\cite{kasai2023evaluating}, strict privacy regulations prevent its use in clinical settings. As a result, research has shifted toward open-source LLMs, but progress is hindered by the scarcity of high-quality Japanese medical data—highlighting the broader challenges faced by low-resource languages.

To address this limitation, in this paper, we are the first to explore the use of Knowledge Graph-based RAG for Japanese medical question answering, specifically targeting small-scale LLMs. Given the limited accessibility of external Japanese medical resources, we use the easily accessible external medical knowledge base UMLS~\cite{bodenreider2004unified} to support knowledge retrieval, following the design of MKG-Rank~\cite{li2025mkg}. By applying word-level translation, LLMs can integrate non-Japanese-centric medical knowledge, mitigating the limitations caused by the relative scarcity of Japanese medical data and its access restrictions. Empirical results show that the KG-based RAG has limited effect on Japanese medical QA with small open-source LLMs, largely depending on the quality and relevance of the retrieved KGs from external knowledge bases. 

\begin{figure*}[t]
    \centering
    \includegraphics[width=1\linewidth]{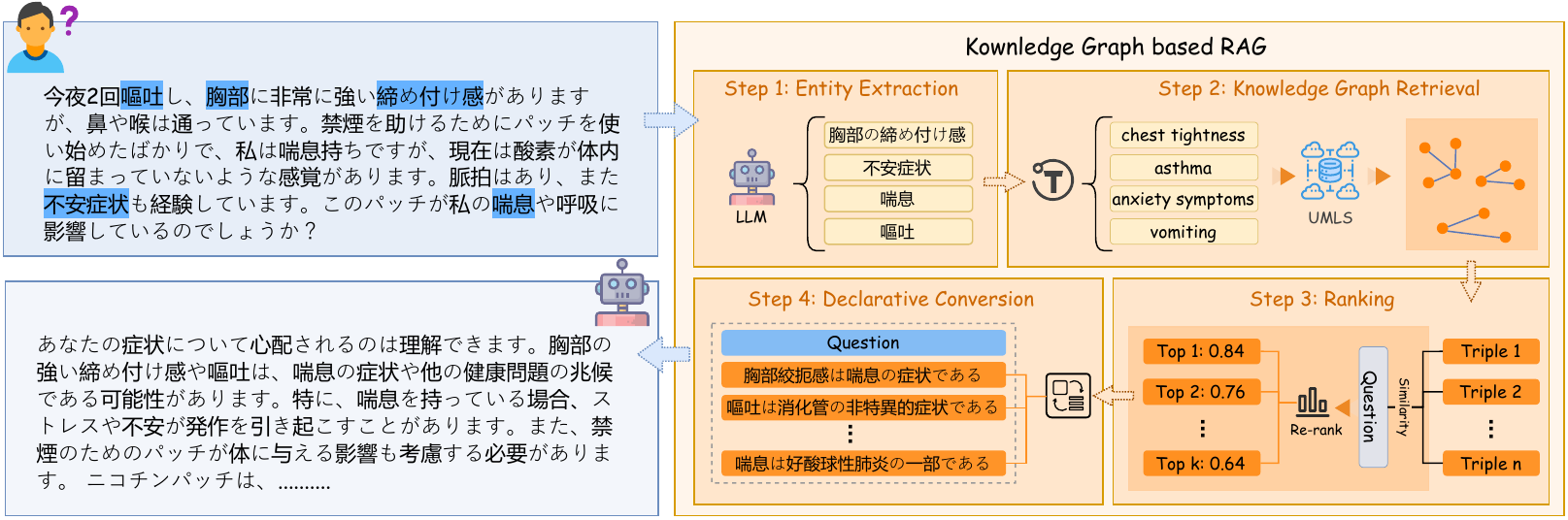}
    \caption{The pipeline of our knowledge graph-based RAG mechanism. Given a Japanese medical question, relevant medical knowledge is retrieved from the external knowledge base and combined with the original Japanese question as input to the LLM for answer generation. The English translation and full content are provided in Appendix~\ref{appendix:pipeline}.}
    \label{fig:pipeline}
    \vspace{-5mm}
\end{figure*}

\vspace{-3mm}
\section{Method}
\vspace{-2mm}

\noindent\textbf{Base Models.} Our work focuses on evaluating the performance of small-scale LLMs in Japanese medical QA. Specifically, we select: (1) models with fewer than 5B parameters, including Borea-Phi-3.5-Mini-Instruct-Common~\cite{Borea-Phi-3.5-mini} and LLaMA-3-ELYZA-JP-8B~\cite{Llama-3-ELYZA-JP-8B}; (2) models with 5B–10B parameters, including LLM-JP-3-7.2B-Instruct3~\cite{llm-jp-3-7.2b-instruct3}, Mistral-7B-Instruct-v0.3~\cite{Mistral-7B-Instruct-v0.3}, Qwen2.5-7B-Instruct~\cite{Qwen2.5-7B-Instruct}, and LLaMA-3.1-8B-Instruct~\cite{Llama-3.1-8B-Instruct}; and (3) 10B–32B models, including Qwen2.5-14B-Instruct~\cite{Qwen2.5-14B-Instruct}, Phi-4-14B~\cite{phi-4}, and Gemma-3-12B-it~\cite{gemma-3-12b-it}. In addition, we evaluate the commercial LLM GPT-4o-mini~\cite{gpt4omini} as a proprietary baseline.


\noindent\textbf{Knowledge Graph-based RAG.} Our pipeline, as illustrated in Fig.~\ref{fig:pipeline}, consists of four main steps: (1) Given a medical question $\mathcal{Q}$, we first use an LLM to extract relevant medical entities $\mathcal{E}$; (2) The extracted entities $\mathcal{E}$ are translated into English via word-level translation and used to query external medical knowledge base UMLS to obtain relevant KGs $\mathcal{G}$. (3) A ranking mechanism is applied to select the most relevant triples $\mathcal{G}^{'}$ based on their semantic relevance to the question $\mathcal{Q}$; (4) The selected triples are then converted into declarative sentences by an LLM, along with the original question, are provided as input to the LLM for answer generation.

\vspace{-3mm}
\section{Experiment}
\vspace{-2mm}

\noindent\textbf{Datasets.}
We selected three long-form medical question answering datasets: ExpertQA-Bio, ExpertQA-Med, and LiveQA. ExpertQA is a high-quality QA dataset verified by domain experts, from which we used 96 biological questions (ExpertQA-Bio) and 504 medical questions (ExpertQA-Med)~\cite{malaviya2023expertqa}. LiveQA consists of consumer health questions submitted to the National Library of Medicine, comprising 627 training QA pairs and 104 test pairs~\cite{abacha2017overview}. All datasets were originally in English and were translated into Japanese using a multi-stage agentic machine translation framework~\cite{xuan2025mmlu}. More details are provided in Appendix~\ref{appendix:dataset}

\begin{table*}[t]
    \centering
    \setlength{\tabcolsep}{2pt}
    \resizebox{1.0\textwidth}{!}{
    \begin{tabular}{lllllll}
         \toprule
         & \multicolumn{2}{c}{ExpertQA-Bio} & \multicolumn{2}{c}{ExpertQA-Med} & \multicolumn{2}{c}{LiveQA} \\
         \cmidrule(lr){2-3} \cmidrule(lr){4-5} \cmidrule(lr){6-7}
         Model & ROUGE-L & BERTScore & ROUGE-L & BERTScore & ROUGE-L & BERTScore\\
         \midrule
         Borea-Phi-3.5~\cite{Borea-Phi-3.5-mini} & 4.33 & 61.20 & 4.92 & 51.54 & 4.05 & 59.98\\
         Borea-Phi-3.5 + RAG & 4.77 (\textcolor[HTML]{009900}{+0.44\%}) & 61.32 (\textcolor[HTML]{009900}{+0.12\%}) & 4.82 (\textcolor[HTML]{CC0000}{-0.10\%}) & 61.39 (\textcolor[HTML]{CC0000}{-0.15\%}) & 3.97 (\textcolor[HTML]{CC0000}{-0.08\%}) & 59.71 (\textcolor[HTML]{CC0000}{-0.28\%})\\
         \midrule
         LLaMA-3-ELYZA-JP-8B~\cite{Llama-3-ELYZA-JP-8B} & 22.29 & 76.51 & 23.12 & 76.81 & 16.23 & 71.51\\
         LLaMA-3-ELYZA-JP-8B + RAG & 22.53 (\textcolor[HTML]{009900}{+0.24\%}) & 76.22 (\textcolor[HTML]{CC0000}{-0.29\%}) & 23.13 (\textcolor[HTML]{009900}{+0.01\%}) & 76.75 (\textcolor[HTML]{CC0000}{-0.06\%}) & 16.37 (\textcolor[HTML]{009900}{+0.14\%}) & 71.56 (\textcolor[HTML]{009900}{+0.05\%})\\
         \midrule
         Mistral-7B~\cite{Mistral-7B-Instruct-v0.3} & 20.85 & 75.25 & 20.21 & 74.67 & 15.70 & 70.99\\
         Mistral-7B + RAG & 17.39 (\textcolor[HTML]{CC0000}{-3.46\%}) & 72.13 (\textcolor[HTML]{CC0000}{-3.12\%}) & 17.59 (\textcolor[HTML]{CC0000}{-2.62\%}) & 72.35 (\textcolor[HTML]{CC0000}{-2.32\%}) & 13.76 (\textcolor[HTML]{CC0000}{-1.94\%}) & 69.15 (\textcolor[HTML]{CC0000}{-1.84\%})\\
         \midrule
         Qwen2.5-7B~\cite{Qwen2.5-7B-Instruct} & 19.09 & 75.15 & 19.52 & 75.39 & 14.60 & 70.88\\
         Qwen2.5-7B + RAG & 20.15 (\textcolor[HTML]{009900}{+1.06\%}) & 75.18 (\textcolor[HTML]{009900}{+0.03\%}) & 20.56 (\textcolor[HTML]{009900}{+1.04\%}) & 74.95 (\textcolor[HTML]{CC0000}{-0.44\%}) & 15.97 (\textcolor[HTML]{009900}{+1.37\%}) & 71.68 (\textcolor[HTML]{009900}{+0.80\%})\\
         \midrule
         LLaMA-3.1-8B~\cite{Llama-3.1-8B-Instruct} & 18.84 & 74.09 & 18.49 & 74.42 & 14.78 & 70.50\\
         LLaMA-3.1-8B + RAG & 18.01 (\textcolor[HTML]{CC0000}{-0.83\%}) & 72.04 (\textcolor[HTML]{CC0000}{-2.05\%}) & 18.01 (\textcolor[HTML]{CC0000}{-0.48\%}) & 72.40 (\textcolor[HTML]{CC0000}{-2.02\%}) & 14.43 (\textcolor[HTML]{CC0000}{-0.35\%}) & 69.30 (\textcolor[HTML]{CC0000}{-1.20\%})\\
         \midrule
         Qwen2.5-14B~\cite{Qwen2.5-14B-Instruct} & 17.00 & 71.92 & 17.01 & 71.44 & 13.85 & 68.63\\
         Qwen2.5-14B + RAG & 17.19 (\textcolor[HTML]{009900}{+0.19\%}) & 71.25 (\textcolor[HTML]{CC0000}{-0.67\%}) & 17.47 (\textcolor[HTML]{009900}{+0.46\%}) & 71.74 (\textcolor[HTML]{009900}{+0.30\%}) & 13.89 (\textcolor[HTML]{009900}{+0.04\%}) & 68.62 (\textcolor[HTML]{CC0000}{-0.01\%})\\
         \midrule
         Phi-4-14B~\cite{phi-4} & 9.72 & 65.89 & 9.15 & 65.12 & 8.78 & 64.97\\
         Phi-4-14B + RAG & 12.18 (\textcolor[HTML]{009900}{+2.46\%}) & 67.38 (\textcolor[HTML]{009900}{+1.49\%}) & 13.93 (\textcolor[HTML]{009900}{+4.78\%}) & 69.13 (\textcolor[HTML]{009900}{+4.01\%}) & 11.18 (\textcolor[HTML]{009900}{+2.40\%}) & 66.75 (\textcolor[HTML]{009900}{+1.78\%})\\
         \midrule
         Gemma-3-12B-it~\cite{gemma-3-12b-it} & 19.79 & 74.65 & 21.25 & 74.89 & 14.61 & 69.71\\
         Gemma-3-12B-it + RAG & 20.44 (\textcolor[HTML]{009900}{+0.65\%}) & 75.13 (\textcolor[HTML]{009900}{+0.48\%}) & 21.99 (\textcolor[HTML]{009900}{+0.74\%}) & 75.71 (\textcolor[HTML]{009900}{+0.82\%}) & 15.34 (\textcolor[HTML]{009900}{+0.73\%}) & 70.42 (\textcolor[HTML]{009900}{+0.71\%})\\
         \midrule
         GPT-4o-mini~\cite{gpt4omini} & 24.92 & 77.67 & 26.78 & 78.67 & 17.94 & 73.25\\
         GPT-4o-mini + RAG & 24.41 (\textcolor[HTML]{CC0000}{-0.51\%}) & 77.45 (\textcolor[HTML]{CC0000}{-0.22\%}) & 26.20 (\textcolor[HTML]{CC0000}{-0.58\%}) & 78.53 (\textcolor[HTML]{CC0000}{-0.14\%}) & 17.85 (\textcolor[HTML]{CC0000}{-0.09\%}) & 73.28 (\textcolor[HTML]{009900}{+0.03\%})\\
         \bottomrule
    \end{tabular}}
    \caption{Evaluation of ROUGE-L and BERTScore for base models and their KG-based RAG-enhanced variants on three medical QA datasets: ExpertQA-Bio, ExpertQA-Med, and LiveQA.}
    \label{tab:main_results}
    \vspace{-8mm}
\end{table*}

\noindent\textbf{Results and Analysis.} Table~\ref{tab:main_results} presents the performance comparison between KG-based RAG and the baseline LLMs on the medical QA task. In general, the results suggest that the impact of our RAG mechanism is relatively limited, with performance fluctuations observed across different base models. Specifically, for Borea-Phi-3.5, LLaMA-3-ELYZA-JP-8B, Qwen2.5-14B, Gemma-3-12B-it, and GPT-4o-mini, the RAG leads to marginal changes, with performance fluctuating within 1\% compared to the respective base models. In contrast, Qwen2.5-7B and Phi-4-14B exhibit overall improvements when enhanced with RAG, particularly Phi-4-14B, which achieves improvements of up to +4.78\% in ROUGE-L and +4.01\% in BERTScore. On the other hand, the RAG leads to negative impacts on Mistral-7B and LLaMA-3.1-8B, resulting in performance degradation.



\begin{CJK}{UTF8}{ipxm}
\begin{table}[t]
    \centering
    \scriptsize
    \resizebox{0.98\textwidth}{!}{
    \begin{tabular}{p{1.0\textwidth}}
    \toprule
        \textbf{Question:}
        \colorbox[HTML]{FFCC99}{ワルファリン}（ワーファリン）を\colorbox[HTML]{FFCC99}{服用}している人は\colorbox[HTML]{FFCC99}{避ける}べき\colorbox[HTML]{FFCC99}{野菜}は何ですか？ (\textcolor[HTML]{004C99}{Which vegetables should people taking warfarin (Coumadin) avoid?})\\
        \textbf{Retrieved Medical Knowledge:}
        1.ワルファリンは生理的に凝固因子濃度を低下させる効果がある。(\textcolor[HTML]{004C99}{Warfarin physiologically reduces the concentration of coagulation factors.}) 2.ワルファリンはクマリン系の抗凝固薬である。(\textcolor[HTML]{004C99}{Warfarin is a coumarin-based anticoagulant.}) \dots \\
        \textbf{Ground Truth:}
        ワーファリンを服用している人は、ビタミンKを多く含む野菜の摂取を避けるか制限する必要があります。なぜなら、それが薬の血液凝固抑制効果（血液をサラサラにする効果）を妨げる可能性があるからです。\dots  (\textcolor[HTML]{004C99}{People taking warfarin should avoid or limit the intake of vegetables that are high in vitamin K, as it may interfere with the medication's blood clotting inhibition effect (the effect of making the blood thinner).}) \\
        \textbf{Generated Answer:}
        ワルファリンを服用している人が避けるべき野菜は、特にビタミンKを多く含むものです。ビタミンKは血液凝固に関与しており、ワルファリンの効果を減少させる可能性があります。 \dots (\textcolor[HTML]{004C99}{People taking warfarin should avoid vegetables that are particularly high in vitamin K. Vitamin K is involved in blood clotting and may reduce the effectiveness of warfarin.})\\
        \midrule
        \textbf{Question:}
        \colorbox[HTML]{FFCC99}{抗生薬}は通常、\colorbox[HTML]{FFCC99}{細菌感染}または\colorbox[HTML]{FFCC99}{真菌感染}のある患者に投与されます。\colorbox[HTML]{FFCC99}{ウイルス}が原因であることが多い疾患については、ではどうでしょうか？ (\textcolor[HTML]{004C99}{Antibiotics are typically administered to patients with bacterial or fungal infections. But what about diseases that are often caused by viruses?})\\
        \textbf{Retrieved Medical Knowledge:}
        1.バイ菌感染は、他のバイ菌性疾患に分類されます。(\textcolor[HTML]{004C99}{Bacterial infections are classified under other bacterial diseases.}) 2.バイ菌感染は、肝膿瘍に分類されます。(\textcolor[HTML]{004C99}{Bacterial infections are classified as liver abscesses.}) \dots \\
        \textbf{Ground Truth:}
        抗生物質は、感冒、インフルエンザ、様々な咳などのウイルス感染症に対しては効果がありません。\dots これは、ウイルスが細胞ではなく、細菌細胞と比較して細胞壁を持たないためです。 \dots (\textcolor[HTML]{004C99}{Antibiotics are ineffective against viral infections such as the common cold, influenza, and various types of cough. \dots This is because viruses do not have a cell wall, unlike bacterial cells. \dots}) \\
        \textbf{Generated Answer:}
        抗生物質は主に細菌感染に対して効果がありますが、ウイルス感染に対しては効果がありません。ウイルスが原因の疾患、例えばインフルエンザや風邪などには、抗生物質は無効です。\dots (\textcolor[HTML]{004C99}{Antibiotics are mainly effective against bacterial infections but are ineffective against viral infections. For illnesses caused by viruses, influenza or the common cold, antibiotics do not work.})\\
    \bottomrule
    \end{tabular}}
    \caption{Case Study. Two cases are presented, with full content and English translations provided in Appendix~\ref{appendix:casestudy}.}
    \vspace{-9mm}
    \label{tab:casestudy}
\end{table}

\noindent\textbf{Case Study.} To further investigate the reasons behind the relatively limited impact of RAG, we analyze the intermediate reasoning process, as shown in Table~\ref{tab:casestudy}.  Specifically, in Case 1 (top), the retrieved medical knowledge includes the information that is relevant and helpful for answering the question (e.g. ``ワルファリンは生理的に凝固因子濃度を低下させる効果がある''). However, we observe that most LLMs already possess this knowledge internally, which explains why RAG offers only marginal improvements in such cases. On the other hand, for Case 2 (bottom), the retrieved medical knowledge does not help answer the question (e.g. ``バイ菌感染は、肝膿瘍に分類されます''). This is primarily because the specialized biomedical concepts in UMLS may not align well with broader, more general medical questions. Consequently, the retrieved information provides limited assistance and may even introduce noise that negatively impacts the LLM’s reasoning.
\end{CJK}

\vspace{-4mm}
\section{Conclusion}
\vspace{-3mm}
This work presents the first exploration of a knowledge graph-based RAG framework for Japanese medical QA using small-scale open-source LLMs. Empirical findings show that its overall impact is limited, primarily constrained by the quality and relevance of the retrieved content. These insights highlight the challenges and potential of applying RAG to Japanese and other low-resource language medical QA tasks.

\section*{Acknowledgment}
This work was supported by JST ACT-X (Grant JPMJAX24CU) and JSPS KAKENHI (Grant 24K20832).


%
%
\bibliographystyle{splncs04}
\bibliography{mybibliography}

\begin{thebibliography}{10}
\providecommand{\url}[1]{\texttt{#1}}
\providecommand{\urlprefix}{URL }
\providecommand{\doi}[1]{https://doi.org/#1}

\bibitem{abacha2017overview}
Abacha, A.B., Agichtein, E., Pinter, Y., Demner-Fushman, D.: Overview of the medical question answering task at trec 2017 liveqa. In: TREC. pp. 1--12 (2017)

\bibitem{Mistral-7B-Instruct-v0.3}
AI, M.: Mistral-7b-instruct-v0.3. https://huggingface.co/mistralai/Mistral-7B-Instruct-v0.3 (2024), retrieved April 08, 2025

\bibitem{Borea-Phi-3.5-mini}
AXCXEPT: Borea-phi-3.5-mini-instruct-common. https://huggingface.co/AXCX EPT/Borea-Phi-3.5-mini-Instruct-Common (2024), retrieved April 08, 2025

\bibitem{bodenreider2004unified}
Bodenreider, O.: The unified medical language system (umls): integrating biomedical terminology. Nucleic acids research  \textbf{32}(suppl\_1),  D267--D270 (2004)

\bibitem{chataigner2024multilingual}
Chataigner, C., Ta{\"\i}k, A., Farnadi, G.: Multilingual hallucination gaps in large language models. arXiv preprint arXiv:2410.18270  (2024)

\bibitem{edge2024local}
Edge, D., Trinh, H., Cheng, N., Bradley, J., Chao, A., Mody, A., Truitt, S., Metropolitansky, D., Ness, R.O., Larson, J.: From local to global: A graph rag approach to query-focused summarization. arXiv preprint arXiv:2404.16130  (2024)

\bibitem{Llama-3-ELYZA-JP-8B}
ELYZA: Llama-3-elyza-jp-8b. https://huggingface.co/elyza/Llama-3-ELYZA-JP-8B (2024), retrieved April 08, 2025

\bibitem{gemma-3-12b-it}
Google: gemma-3-12b-it. https://huggingface.co/google/gemma-3-12b-it (2025), retrieved April 08, 2025

\bibitem{hurst2024gpt}
Hurst, A., Lerer, A., Goucher, A.P., Perelman, A., Ramesh, A., Clark, A., Ostrow, A., Welihinda, A., Hayes, A., Radford, A., et~al.: Gpt-4o system card. arXiv preprint arXiv:2410.21276  (2024)

\bibitem{kasai2023evaluating}
Kasai, J., Kasai, Y., Sakaguchi, K., Yamada, Y., Radev, D.: Evaluating gpt-4 and chatgpt on japanese medical licensing examinations. arXiv preprint arXiv:2303.18027  (2023)

\bibitem{ke2024mitigating}
Ke, Y., Yang, R., Lie, S.A., Lim, T.X.Y., Ning, Y., Li, I., Abdullah, H.R., Ting, D.S.W., Liu, N.: Mitigating cognitive biases in clinical decision-making through multi-agent conversations using large language models: simulation study. Journal of Medical Internet Research  \textbf{26},  e59439 (2024)

\bibitem{li2025mkg}
Li, F., Chen, Y., Liu, H., Yang, R., Yuan, H., Jiang, Y., Li, T., Taylor, E.M., Rouhizadeh, H., Iwasawa, Y., et~al.: Mkg-rank: Enhancing large language models with knowledge graph for multilingual medical question answering. arXiv preprint arXiv:2503.16131  (2025)

\bibitem{lin2004rouge}
Lin, C.Y.: Rouge: A package for automatic evaluation of summaries. In: Text summarization branches out. pp. 74--81 (2004)

\bibitem{Llama-3.1-8B-Instruct}
Llama, M.: Llama-3.1-8b-instruct. https://huggingface.co/meta-llama/Llama-3.1-8B-Instruct (2024), retrieved April 08, 2025

\bibitem{llm-jp-3-7.2b-instruct3}
LLM-jp: llm-jp-3-7.2b-instruct3. https://huggingface.co/llm-jp/llm-jp-3-7.2b-instruct3 (2024), retrieved April 08, 2025

\bibitem{malaviya2023expertqa}
Malaviya, C., Lee, S., Chen, S., Sieber, E., Yatskar, M., Roth, D.: Expertqa: Expert-curated questions and attributed answers. arXiv preprint arXiv:2309.07852  (2023)

\bibitem{mckenna2023sources}
McKenna, N., Li, T., Cheng, L., Hosseini, M.J., Johnson, M., Steedman, M.: Sources of hallucination by large language models on inference tasks. arXiv preprint arXiv:2305.14552  (2023)

\bibitem{phi-4}
Microsoft: phi-4. https://huggingface.co/microsoft/phi-4 (2024), retrieved April 08, 2025

\bibitem{nori2023capabilities}
Nori, H., King, N., McKinney, S.M., Carignan, D., Horvitz, E.: Capabilities of gpt-4 on medical challenge problems. arXiv preprint arXiv:2303.13375  (2023)

\bibitem{gpt4omini}
OpenAI: Gpt-4o-mini. https://openai.com/index/

\bibitem{Qwen2.5-14B-Instruct}
Qwen: Qwen2.5-14b-instruct. https://huggingface.co/Qwen/Qwen2.5-14B-Instruct (2024), retrieved April 08, 2025

\bibitem{Qwen2.5-7B-Instruct}
Qwen: Qwen2.5-7b-instruct. https://huggingface.co/Qwen/Qwen2.5-7B-Instruct (2024), retrieved April 08, 2025

\bibitem{shi2023mkrag}
Shi, Y., Xu, S., Yang, T., Liu, Z., Liu, T., Li, Q., Li, X., Liu, N.: Mkrag: Medical knowledge retrieval augmented generation for medical question answering. arXiv preprint arXiv:2309.16035  (2023)

\bibitem{touvron2023llama}
Touvron, H., Lavril, T., Izacard, G., Martinet, X., Lachaux, M.A., Lacroix, T., Rozi{\`e}re, B., Goyal, N., Hambro, E., Azhar, F., et~al.: Llama: Open and efficient foundation language models. arXiv preprint arXiv:2302.13971  (2023)

\bibitem{xiong-etal-2024-benchmarking}
Xiong, G., Jin, Q., Lu, Z., Zhang, A.: Benchmarking retrieval-augmented generation for medicine. In: Ku, L.W., Martins, A., Srikumar, V. (eds.) Findings of the Association for Computational Linguistics: ACL 2024. pp. 6233--6251. Association for Computational Linguistics, Bangkok, Thailand (Aug 2024). \doi{10.18653/v1/2024.findings-acl.372}, \url{https://aclanthology.org/2024.findings-acl.372/}

\bibitem{xuan2025mmlu}
Xuan, W., Yang, R., Qi, H., Zeng, Q., Xiao, Y., Xing, Y., Wang, J., Li, H., Li, X., Yu, K., et~al.: Mmlu-prox: A multilingual benchmark for advanced large language model evaluation. arXiv preprint arXiv:2503.10497  (2025)

\bibitem{yang2024llm}
Yang, H., Chen, H., Guo, H., Chen, Y., Lin, C.S., Hu, S., Hu, J., Wu, X., Wang, X.: Llm-medqa: Enhancing medical question answering through case studies in large language models. arXiv preprint arXiv:2501.05464  (2024)

\bibitem{yang2024kg}
Yang, R., Liu, H., Marrese-Taylor, E., Zeng, Q., Ke, Y.H., Li, W., Cheng, L., Chen, Q., Caverlee, J., Matsuo, Y., et~al.: Kg-rank: Enhancing large language models for medical qa with knowledge graphs and ranking techniques. arXiv preprint arXiv:2403.05881  (2024)

\bibitem{yang2025retrieval}
Yang, R., Ning, Y., Keppo, E., Liu, M., Hong, C., Bitterman, D.S., Ong, J.C.L., Ting, D.S.W., Liu, N.: Retrieval-augmented generation for generative artificial intelligence in health care. npj Health Systems  \textbf{2}(1), ~2 (2025)

\bibitem{zhang2023language}
Zhang, M., Press, O., Merrill, W., Liu, A., Smith, N.A.: How language model hallucinations can snowball. arXiv preprint arXiv:2305.13534  (2023)

\bibitem{zhang2019bertscore}
Zhang, T., Kishore, V., Wu, F., Weinberger, K.Q., Artzi, Y.: Bertscore: Evaluating text generation with bert. arXiv preprint arXiv:1904.09675  (2019)

\end{thebibliography}
%






\appendix
\section{Dataset Details}
\label{appendix:dataset}
We provide detailed statistics of the evaluation datasets, including the number of samples and the average word count of questions and answers, as shown in Table~\ref{tab:dataset}.

\begin{table}[h]
  \centering
  \setlength{\tabcolsep}{6pt}
  \resizebox{0.8\textwidth}{!}{
  \begin{tabular}{lccc} 
    \toprule
    Dataset & Size & Question Length &  Answer Length \\
    \midrule
    ExpertQA-Bio & 96 & 56.7 & 410.7 \\
    ExpertQA-Med & 504 & 56.0 & 378.1 \\
    LiveQA & 627 & 118.9 & 438.3 \\
    \bottomrule
  \end{tabular}}
  \caption{Statistics of the evaluation datasets.}
  \label{tab:dataset}
\end{table}

\section{Metrics}
We adopt ROUGE-L~\cite{lin2004rouge} and BERTScore~\cite{zhang2019bertscore} to evaluate the quality of answers generated by LLMs for Japanese medical QA tasks. These metrics provide a comprehensive assessment of both lexical and semantic accuracy.

\section{English Translation of the Content in Fig.~\ref{fig:pipeline}}
\label{appendix:pipeline}
\begin{figure*}[t]
    \centering
    \includegraphics[width=1\linewidth]{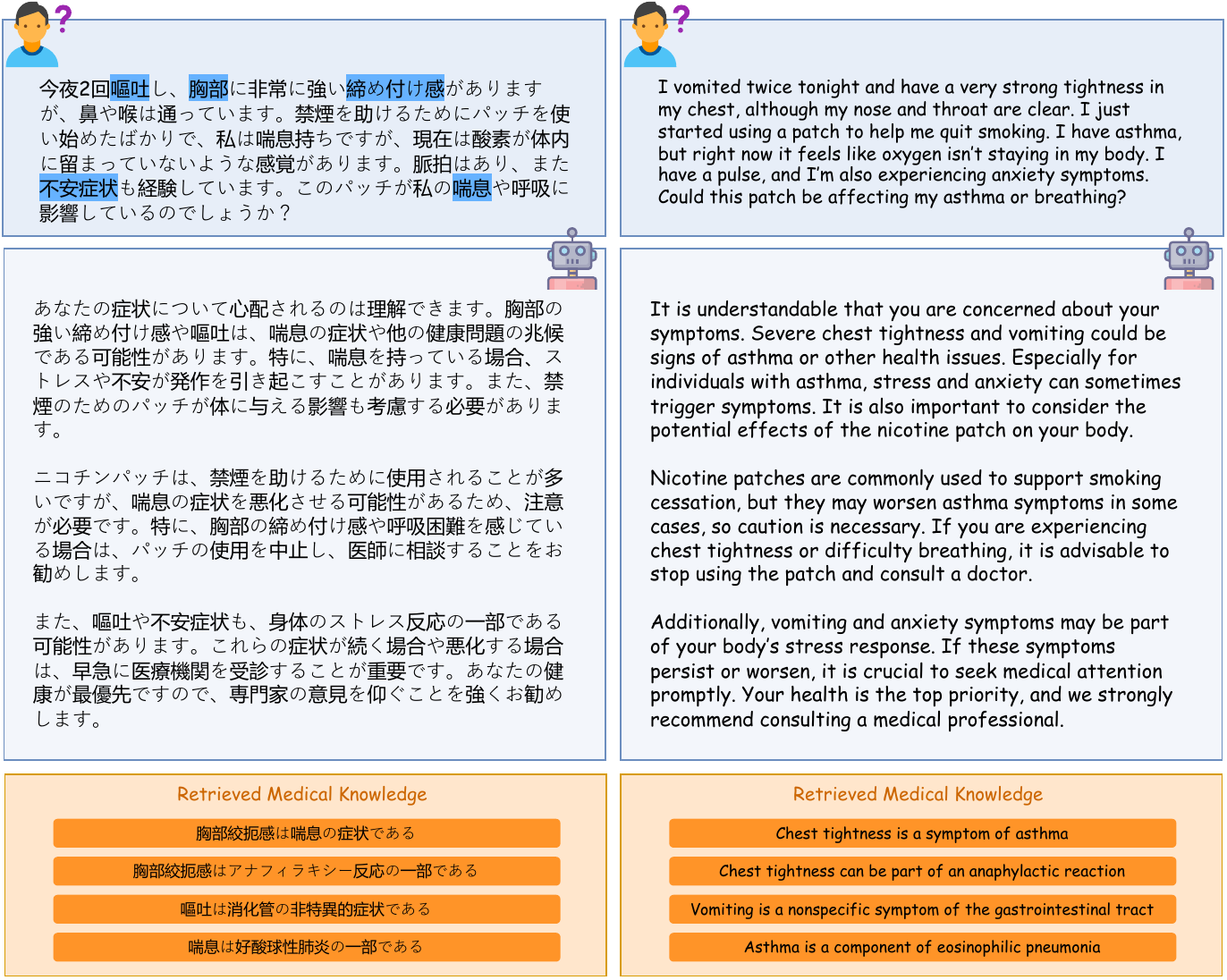}
    \caption{English translation and the full content of the question and answer in the Fig.~\ref{fig:pipeline}}
    \label{fig:appendix_pipeline}
\end{figure*}

For clearer presentation, we provide the English translation and the full content of the question and answer in the Fig.~\ref{fig:pipeline}, as illustrated in Fig.~\ref{fig:appendix_pipeline}.

\section{prompts}
In this section, we provide the prompts used in our knowledge graph-based RAG mechanism, including Medical Entity Extraction (Table~\ref{tab:prompt1}), Declarative Conversion (Table~\ref{tab:prompt2}) and Answer Generation (Table~\ref{tab:prompt3}).

\begin{table}[h]
    \centering
    \resizebox{1.0\textwidth}{!}{
    \begin{tabular}{p{1.0\textwidth}}
    \toprule
        Medical Entity Extraction \\
        \midrule
        text: \{\textbf{question}\} \\
        Please extract at most 4 terms related to medical that you think are the most important from the provided text.  \\            
        Returns the result in the following json form. All the results are merged into one json.  \\
        \texttt{--} Examples of results: \\
        \{"medical terminologies" : ["term1", "term2", ...]\} \\
        \\
        result: \\
    \bottomrule
    \end{tabular}}
    \caption{Prompt Used for Medical Entity Extraction.}
    \label{tab:prompt1}
\end{table}

\begin{CJK}{UTF8}{ipxm}
\begin{table}[h]
    \centering
    \resizebox{1.0\textwidth}{!}{
    \begin{tabular}{p{1.0\textwidth}}
    \toprule
        Declarative Conversion \\
        \midrule
        あなたは医学分野の知能助手です。 \\
        すべての背景知識をそれぞれ日本語の平叙文に変換する。 医学に関係ないと思うものは何でも削除できます。  \\   
        - Background Knowledge: \{\textbf{triple}\}  \\
        \\
        Converted Background Knowledge: \\
    \bottomrule
    \end{tabular}}
    \caption{Prompt Used for Declarative Conversion.}
    \label{tab:prompt2}
\end{table}
\end{CJK}

\section{Case Study Details}
\label{appendix:casestudy}
\begin{figure*}[t]
    \centering
    \includegraphics[width=1\linewidth]{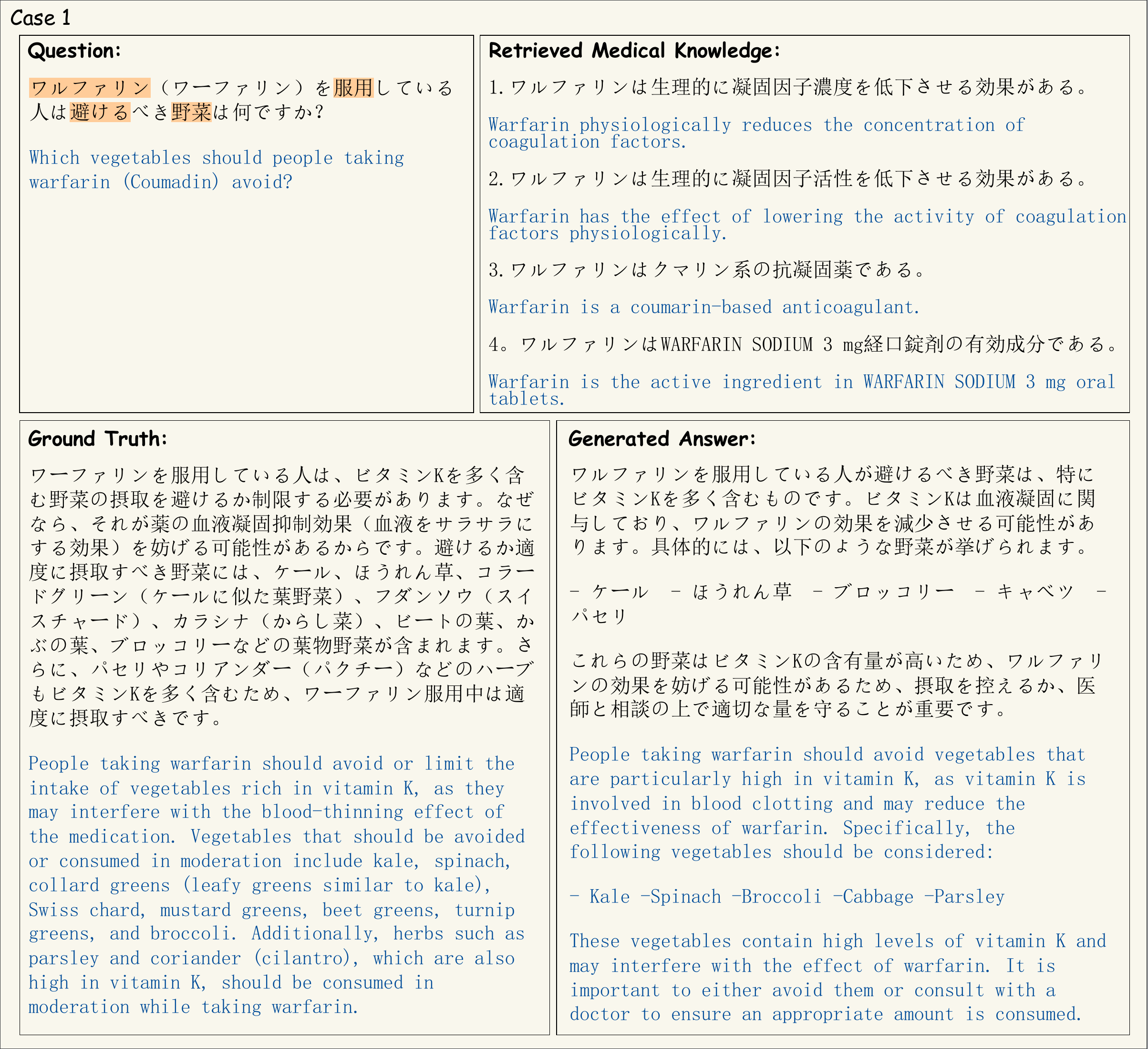}
    \caption{Full content and English translation of the case 1 in Table~\ref{tab:casestudy}.}
    \label{fig:appendix_casestudy1}
\end{figure*}

\begin{figure*}[t]
    \centering
    \includegraphics[width=1\linewidth]{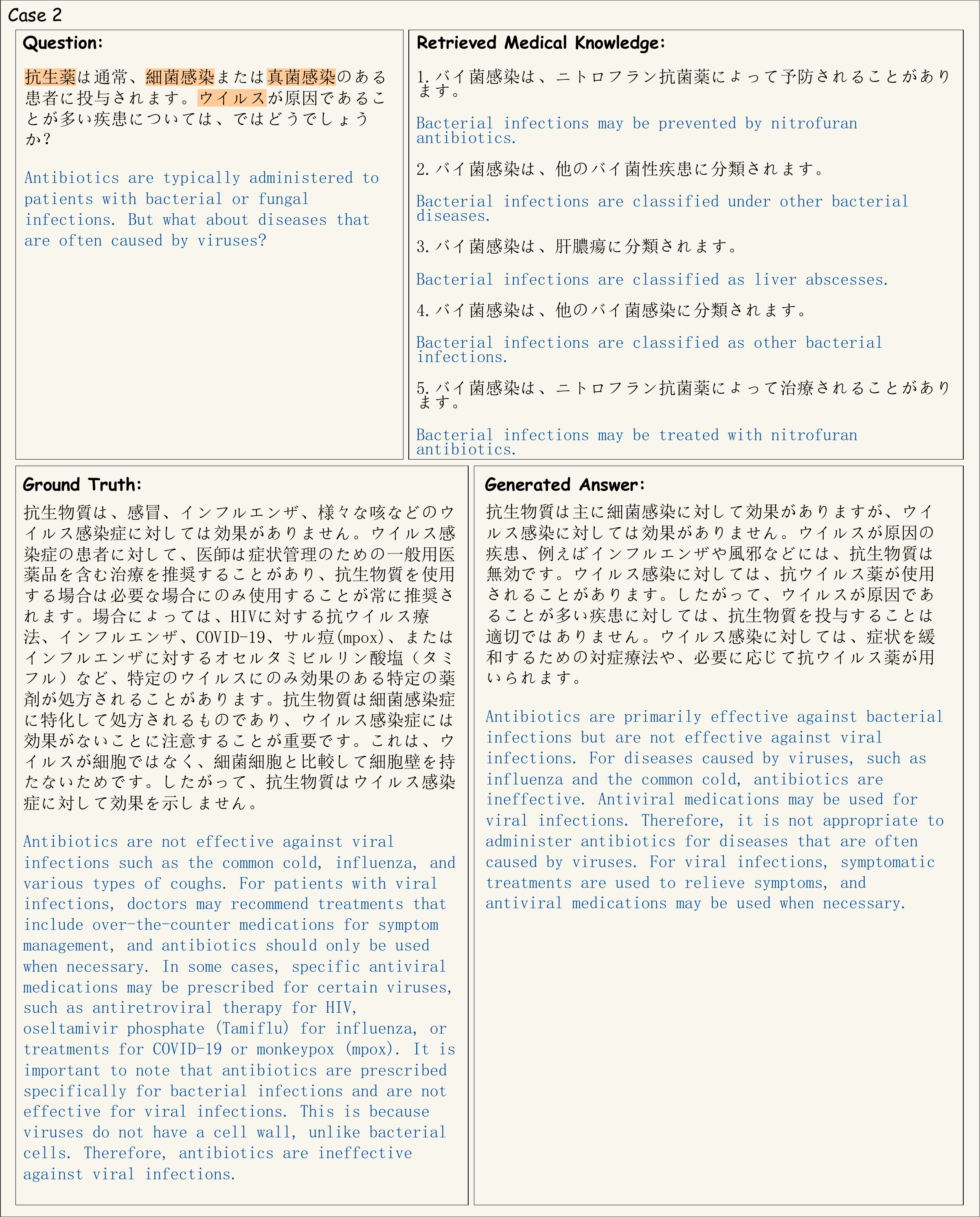}
    \caption{Full content and English translation of the case 2 in Table~\ref{tab:casestudy}.}
    \label{fig:appendix_casestudy2}
\end{figure*}

For clarity and ease of understanding, we provide the full content and English translations of the cases from Table~\ref{tab:casestudy} here, as shown in Fig.~\ref{fig:appendix_casestudy1} and Fig.~\ref{fig:appendix_casestudy2}.

\begin{CJK}{UTF8}{ipxm}
\begin{table}[h]
    \centering
    \resizebox{1.0\textwidth}{!}{
    \begin{tabular}{p{1.0\textwidth}}
    \toprule
        Answer Generation with Medical Knowledge \\
        \midrule
        あなたは医学分野の知能助手です。 質問をよく分析し、提供された背景知識とあなた自身の知識に基づいて以下の質問に答えてください。 できるだけ512のtoken内で完全に回答します。 \\
        日本語で質問に答える。  \\   
        \\
        - 問題: \{\textbf{question}\}  \\
        - 背景知識: \{\textbf{background\_knowledge}\}  \\
        \\
        - 答える: \\
    \bottomrule
    \end{tabular}}
    \caption{Prompt Used for Answer Generation with Medical Knowledge.}
    \label{tab:prompt3}
\end{table}
\end{CJK}

\end{document}